# BUILDING A SYLLABLE DATABASE TO SOLVE THE PROBLEM OF KHMER WORD SEGMENTATION

Tran Van Nam[1], Nguyen Thi Hue[2] and Phan Huy Khanh[3]

[1]Department of Computer Engineering; University of Tra Vinh, Vietnam
[2]School of Southern Khmer Language; University of Tra Vinh, Vietnam
[3]Department of Computer Engineering; Polytechnic University of Da Nang, Vietnam

## ABSTRACT

*Word segmentation is a basic problem in natural language processing. With the languages having the complex writing system like the Khmer language in Southern of Vietnam, this problem really very intractable, posing the significant challenges. Although there are some experts in Vietnam as well as international having deeply researched this problem, there are still no reasonable results meeting the demand, in particular, no treated thoroughly the ambiguous phenomenon, in the process of Khmer language processing so far. This paper present a solution based on the syllable division into component clusters using two syllable models proposed, thereby building a Khmer syllable database, is still not actually available. This method using a lexical database updated from the online Khmer dictionaries and some supported dictionaries serving role of training data and complementary linguistic characteristics. Each component cluster is labelled and located by the first and last letter to identify entirety a syllable. This approach is workable and the test results achieve high accuracy, eliminate the ambiguity, contribute to solving the problem of word segmentation and applying efficiency in Khmer language processing.*

## 1. INTRODUCTION

In natural language processing, the problem of word segmentation (WS), or dividing a string of written language into its component words in a document, always put out immediately after many processing steps such as encoding, building typing tool, editing documents, etc. For the languages using Latin alphabet such as English, Vietnamese, or many other languages in the world, the problem of WS does not pose much difficulty. In such writing systems, there are the delimiters (spaces, or punctuation) between words to identify exactly any word appeared in document. The general characteristics of WS in these languages are identified by their meanings and eliminate ambiguity, depending on context of documents. Several methods have been proposed and familiar as the maximum matching method, finding the conditional random field, learning machine, support vector machines, hidden Markov models and so on [4][6][8][9].

The Khmer language belongs to South Asian group (Thai, Laotian, Vietnamese…) having a very complicated writing system. The letters are writing seamless and in order from top to bottom, left to right, but not using any delimiter between words. As written, the letters or signs can be placed below, above, front and behind a central character [1]. A word can be read in many different ways depending on the context of document. There are many ways to syllable formation, causing the ambiguity in orthography and meanings [1][2][3][24]. Therefore, all language processing approaches on the computer, especially the problem of WS, always deal with a big challenge, it has not been resolved in a satisfactory manner so far.

The WS methods for Khmer documents is mainly based on two approaches: on the characters and on the words. The first approach carried out by groups of characters respectively appearing in sentences into syllables, depending on the case where the characters can repeated. For example the string ABCD can be split into AB CD or AB BC CD, and so on... The second approach seeks to separate complete words (single or compound) in a sentence in three ways: basing on statistical frequency of the words, using existing vocabulary in a lexical database (LDB), or the combination of these both ways [6][7][10][11]. However, the effectiveness of this methods is still not high, it should be studied further. The drawback in general is unable to recognize the new words as well as simple word or compound words, it does not resolve the ambiguity phenomenon. On the other hand, these methods are not yet fully exploited these specific characteristics of Khmer language in the Southern Vietnam, it has changed many factors, borrowed words become more complex than original Cambodia's Khmer language.

In approach based on the words, an other method proposed is building graphs analyzing sentences [8]. This method identifies new words by searching, analyzing lists which may correspond to the shortest paths on a graph [7]. Although the WS is relatively effective, can be resolve ambiguity problem, but this method is only suitable for languages having the signs of separation or identification for syllables and using a syllable database (SDB). However, there is not yet SDB for Khmer language built so far using for universal services for many different purposes, such as spelling checking, classification of Khmer documents, etc.

This paper proposed a solution for building a Khmer SDB. After the introduction, the paper mainly includes the following content: analysing all features of Khmer language; proposing two syllable models, SC and CC, to recognize the characters combined in component clusters of a syllable; dividing each entry from a LDB into component clusters; labeling specific position based on the first and last letters of each cluster to determine the limits of each syllable, updating processed syllables in the SDB; test implementation and evaluation of the results. The last part of paper is the conclusion and directions for further research.

## 2. CHARACTERISTICS OF KHMER WRITING SYSTEM

### 2.1. An introduction to Khmer language

Khmer (កាសាខ្មែរ [pʰiːɔsa: kʰmaːe]) is the official language of Cambodia. The Khmer language belongs to 150 Mon-Khmer language in Southeast Asia, mainly distributed in Cambodia, Thailand. In Vietnam, besides Khmer people living in Southern Vietnam, there are 19 ethnic groups who have the same language MonKhmer, scattered mainly in the Central Highlands like Bana, XeDang, Koho, Hre, Mnong, Stieng, Cotu, etc. These languages are so far not virtually researched in the domain of natural language processing The Khmer language is monosyllable, voiceless (no dialcritical marks like Vietnamese, Laotian and Thai), borrowed from many languages, including Sanskrit, Thai, Cham, Chinese, Vietnamese, including France, Portugal [2][5]. In Vietnam, Young Khmer generation has borrowed many Vietnamese words. For example, in the sentence គេធ្វើបវសាក់អ៊ីនិង? (*What do they announce?*), the new word ផ្សព្វ (*announce*) is borrowed from Vietnamese language. The Khmer grammar uses word family, sentence structure, sentence classe et types, etc. but it is not too complicated.

### 2.2. Khmer alphabet

The Khmer alphabet has over 120 characters including consonants, vowels and signs. The TABLE I. introduce 33 consonants: 15 consonants voiced O [ɔ] and 18 consonants voiced Ô [o]. Only the consonant ឫ in voice O has not the leg sign, the 32 remaining consonants all have leg signs, dominant the syllable formation [1][3][7][14].

TABLE I. THE CONSONANTS.

| Consonants voiced O [ɔ] | | | | | | | | |
|---|---|---|---|---|---|---|---|---|
| ក | ខ | គ | ឃ | ង | ច | ឆ | ជ | ឈ |
| ញ | ដ | ឋ | ឌ | ឍ | ណ | | | |

| Consonants voiced Ô [o] | | | | | | | | |
|---|---|---|---|---|---|---|---|---|
| ត | ថ | ទ | ធ | ន | ប | ផ | ព | ភ |
| ម | យ | រ | ល | វ | ស | ហ | ឡ | អ |

In our research, we used conventionally the leg sign ្ preceding the consonant having the leg so that the computers can recognize them in consonant combinations. It is true of the consonant combination ្រ having a ្ before the consonant រ.

TABLE II. THE VOWELS.

| Dependent Vowels | | | | | | | | | | | | | |
|---|---|---|---|---|---|---|---|---|---|---|---|---|---|
| ា | ិ | ី | ឹ | ឺ | ុ | ូ | ួ | ើ | ឿ | ៀ | េ | ែ | ៃ |
| ោ | ៅ | ុំ | ំ | ាំ | ះ | ុះ | េះ | ោះ | ុំះ | | | | |

| Independent vowels | | | | | | | | | | | | | |
|---|---|---|---|---|---|---|---|---|---|---|---|---|---|
| ឥ | ឦ | ឧ | ឩ | ឪ | ឫ | ឬ | ឭ | ឮ | ឯ | ឰ | ឱ | ឲ | ឳ |

There are 24 common vowels (or dependent) and 14 independent vowels. A common vowel always has a meaning even when it is not combined with any other consonants, for example, ឦ (*northeast*), ឥ (*rainbow*), etc. An independent vowel only has a meaning when it is combined with other consonants, the pronunciation is depending on the consonants in voice O or in voice Ô.

The TABLE III. below introduce a list of signs and 10 digits from 0 to 9. In the numeral system, the reading 6 7 8 9 is respectively reassembling 5 with 1 2 3 4. The numbers of tens onwards normally use the familiar numeral system, for example ១០ (10), ១៩ (19), and so on.

TABLE III. THE SIGNS AND DIGITS.

| Three types of signs | | | | | |
|---|---|---|---|---|---|
| Signs at the begin of syllable | ៏ | ៌ | ៍ | ៎ | ៏ |
| Signs at the end of syllable | ៉ | ៊ | ះ | ៗ | |
| Signs of replacement for consonant រ or overlapped consonants | ៌ | | | | |
| The other signs | | | | | |
| ។ | ៕ | ៙ | ? | ៘ | ៚ | ! | = | « | » |
| . | ( | ) | , | @ | ៌ | " | $ | ¢ | € |
| % | * | \| | / | / | x | | | | |
| The digits | | | | | |
| ០ | ១ | ២ | ៣ | ៤ | ៥ | ៦ | ៧ | ៨ | ៩ |

The Khmer writing system uses a structure of three levels (similar to Thai and Laotian): leg (*below*), body and hair (*above*). The leg includes the vowels, the body include the combination of consonants and vowels, the hair include the vowels or signs. For example, the word មេ (*me*) consists of three levels: the leg is the vowel ្យ, the body is the combination of consonant ម and the hair is the vowel េ.

## 2.3. Khmer syllable formation

The syllables in Khmer are used to form the words. In reality, there are three types of single word formation: semi-syllable as ខ្ញុំ (*I*), one syllable as កង់ (*bicycle*), or two syllables as ចេកគ្រៀប (*banana bunch*). A compound word has at least two single words. In case of compound word with two syllable, just one of the two syllables has meaning. For example, កលិន (*idiom*) has only one syllable (*pouring water*) has meaning. A phrase consists of a group of words having at least three syllables.

The Khmer syllable formation is very complex. There are four ways for combinations into syllables: a consonant is combined with a other consonant as បេ (*he*); a consonant is combined with a vowel as ជី (*fertilizer*); a consonant coupling with a vowel is combined with a other consonant as ស្តីុ (*study*); a consonant having the leg sign is combined with a vowel as ស្រែ (*field*).

The dependent vowels in the TABLE II are always behind consonants, though there are possible 9 different positions: front, back, top, bottom, front back, on the bottom, front on, on the back and lower back. When there are two syllables combined, the final consonant of the first syllable is written overlapping on the first consonant of the second syllable. For example, in កណ្តាស់ (sneezing), the final consonant ណ of the syllable កណ is written overlapping on the initial consonant ត of the syllable តាស់. So, the Khmer syllable formation often causes ambiguity. It is true that there are two ways of divide the word ទារក (*infant*) into two syllables, or ទា | រក (*duck/ find*), or ទារ | ក (*hungry/ neck*).

## 3. BUILDING THE SYLLABLE DATABASE

### 3.1. Proposing the Khmer syllabble models

Basing on the specific characteristics in Khmer writing system, we propose two syllabble models: simple combination (SC) and cluster combination (CC). So, we use the following convention:

- [X] means character X may be absent (Option).
- (X) means character X is at the beginning position of cluster when it is in the same group Y or Z.

### 3.1.1. SC, the simple combination to syllable formation

The SC model determine the simple syllable combination containing maximum two consecutive letters. Any syllable in this model has the form **V[C]**, where **V** is a independent vowel, **C** is a consonant. The TABLE IV. below illustrates the syllables formed by one or two letters:

TABLE IV.    SYLLABLES FORMATION USING SC

| Syllable | V | C |
|---|---|---|
| ឥ | ឥ | |
| ឪ | ឪ | |
| ឫស | ឫ | ស |
| ឥក | ឥ | ក |

### 3.1.2. CC, the complex combination to syllable formation

The CC model determine the syllable combination in the form of three component clusters placed behind a consonant **C**<Begin><Center><End>:

<Begin> = **C**$[X_1][X_2][X_3][X_4]$
<Center>     = $(Y_5)[Y_6](Y_7)[Y_8][Y_9]$
<End>     = $(Z_{10})[Z_{11}][Z_{12}]$

Where:

- **C**, $Y_5$, $Z_{10}$, $Z_{11}$ are consonants.
- $X_2$, $X_3$, $Y_7$, $Y_8$ are consonants with leg signs.
- $X_4$, $Y_9$ dependent vowel.
- $X_1$ is the beginning sign of syllable.
- $Y_6$ is the sign ្ or the overlapping consonant.
- $Z_{12}$ is the common sign ending syllable.
-

The TABLE V. below illustrates some typical cases how to identify the characters appeared in a syllable according to CC model:

TABLE V.    SYLLABLE IN CC MODEL.

| Syllable | IPA | C | $X_1$ | $X_2$ | $X_3$ | $X_4$ | $Y_5$ | $Y_6$ | $Y_7$ | $Y_8$ | $Y_9$ | $Z_{10}$ | $Z_{11}$ | $Z_{12}$ |
|---|---|---|---|---|---|---|---|---|---|---|---|---|---|---|
| ក | kɑ | ក | | | | | | | | | | | | |
| ម៉ែ | maːe | ម | ៉ | | | ែ | | | | | | | | |
| ខ្មែ | kʰmaːe | ខ | | ្ម | | ែ | | | | | | | ៎ | |
| កា | kaː | ក | | | | | | | | | | | | |
| ហ្គ្រាយ | hkraːiː | ហ | | ្គ | ្រ | ា | | | | | | យ | | |
| អង្គ្រង | aŋ-krɔŋ | អ | | | | ង | | | ្គ | ្រ | | ង | | |
| អុច | oc-oc | អ | | | | ុ | | | | | | ច | | ៗ |
| កង្វា | kaŋ-ʋaː | ក | | | | ង | | | ្វ | | ា | ៎ | | |
| មិអកិអ | miːə-kiːə | ម | | | | ា | ក | ៝ | | | ា | | | |
| ប៉ង | ɓaŋ | ប | | | | | | | | | | ង | | |
| កាណ៌ | kaː | ក | | | | ា | | | | | | ៎ | ណ៌ | ៝ |

Applying the CC model, the TABLE VI. illustrate the dividing of some entries of LDB into component clusters which placed in brackets []. The longest syllable containing completly three clusters, between two component clusters separated by sign +.

TABLE VI.    SYLLABLE FORMATION USING CC.

| Entry | Type | Meaning | Clusters | Syllables |
|-------|------|---------|----------|-----------|
| និង | Simple word | *and/ with* | [និ]+[ង] | និ\|ង |
| បង្រៀន | Simple word | *teachers/ teaching* | [ប]+[ង្រៀ]+[ន] | ប\|ង្រៀ\|ន |
| កប៉ាល់ | Compound word | *ship* | [ក]+[ប៉ា]+[ល់] | ក\|ប៉ា\|ល់ |
| ពូរុកអ៊ំ | Phrase | *uncle* | [ពូ]+[រុ]+[ក]+[អ៊ំ] | ពូ\|រុ\|ក\|អ៊ំ |
| បរិយាកាស | Phrase | *atmosphere/ air* | [ប]+[រិ]+[យា]+[ក]+[ស] | ប\|រិ\|យា\|ក\|ស |

With these proposed models, separated clusters are the smallest ones with leg signs, depenent vowel cannot stand by itself, this violates Khmer syllabic structure.

### 3.1.3. Applying Khmer syllable models

Both two models SC and CC uses the characcteristic signs to identify any component clusters that is the begin, the center or the end of a syllable.

TABLE VII.    BEGIN OF SYLLABLE.

| Dependent vowel | កា | ន | ន | ប្ប | ប្ប | ថ | ប្ប | ថ |  |
|-----------------|-----|----|----|-----|-----|----|-----|----|----|
| Beginning consonant | ន | ណា | ម | ន | ណា | ស | ហា | ក្ក | អ |
| Having signs | ◌៉ | ◌៊ | ◌់ |  |  |  |  |  |  |
| Simple syllable and dependent vowel |  |  |  |  |  |  |  |  |  |
|  |  |  |  |  |  |  |  |  |  |

TABLE VIII.    END OF SYLLABLE.

| Double vowels | ◌ៈ | ◌ំ | ◌ុំ | ◌ះ | ◌ោះ | ◌ោ | ◌ុំ | ◌ៃៈ |
|---------------|-----|-----|------|-----|------|------|------|------|
| Having signs | ◌់ | ◌ះ | ◌់ | ◌ង |  |  |  |  |
|  |  |  |  |  |  |  |  |  |

TABLE IX.    THE BEGIN IS AS THE END OF SYLLABLE.

| Having characters |  |  | ◌់ | ◌់ |  |  |  |  |
|-------------------|--|--|-----|-----|--|--|--|--|
| Double-independent vowel | ឆ្លូ | ឥ | ឫ | ឩ | ឯ | ឰ |  |  |
| Numbers, Latin characters, special characters, or borrowed character |  |  |  |  |  |  |  |  |

## 3.2. Building the syllable database

0illustrates the solution of using two models SC and CC to build the Khmer SDB. The solutions use 7 **databases (DB)** from (1) to (7). The input (1) is a LDB containing the Khmer vocabularies in social sector, no containing or borrowed words or other non-textual components. The Unicode and DaunPenh fonts are used because of their popularity and available for use at present. This LDB contained 48,947 vocabulary entries which are updated from some online Khmer dictionaries [13][14]. The entries are arranged in the dictionary order (ABC) which can be single words, compound words, or phrases.

The solutions use is summarized into five following steps:

### 3.2.1. Dividing into component clusters

Using the models SC or CC, this step divided each entry of LDB into syllables by identifying each its character being at the begin (V or C), the center (Y5), or the end position (Z10) of a appeared components cluster which can be formed a new syllable. For example, the entry វិញ្ញាសាស្រ្តកុំព្យូទ័រ (*Information Technology*) is divided into 8 component clusters: [វិ]+[ញ្ញា]+[សា]+[ស្រ្ត]+[កុំ]+[ព្យូ]+[ទ័]+[រ].

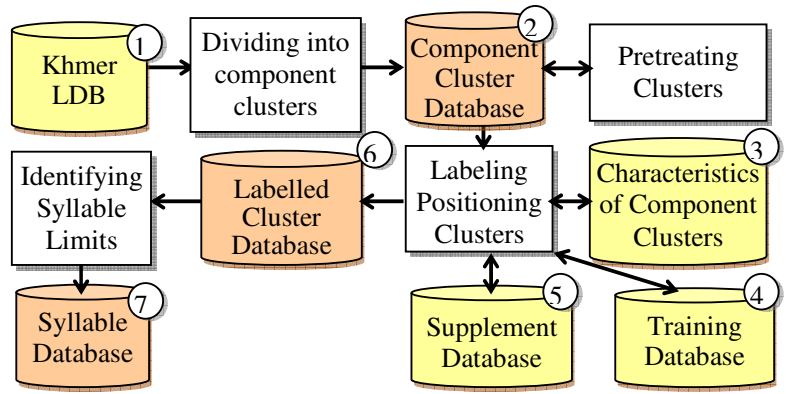

Figure 1. Building Khmer syllable database.

### 3.2.2. Data preprocessing

To properly identify syllables Khmer, the pre - processing step removes the misspelled characters and rearrange the consistent order of characters which appeared in each component cluster based on the models SC or CC. This step is based on three phases:
*Handling the position of each character in syllables*

This phase is related to the way how to entry Khmer characters into the computer. At present, almost editor system for Khmer documents use Unicode and is available on the internet, or installed by default in Windows. The typing method for Khmer documents is similar with Vietnamese documents from a ASCII keyboard. The users typed in a row any character forming the words and sentences of document, in order from top to bottom, left to right. Usually, there are two typing methods: normal typing and NIDA[1] typing [12]. A Khmer syllable can be typing in different ways, each way forming a combination of independent characters. For example, there are 4 different combinations of syllable ស្រ្តី (women), each combination containing 6 characters to be typed in turn as follow:

(1) ស + ◌ុ + ន + ◌ុ + ɪ + ◌ី = (ស, ◌ុ, ន, ◌ុ, ɪ, ◌ី)

(2) ស + ◌ុ + ɪ + ◌ុ + ន + ◌ី = (ស, ◌ុ, ɪ, ◌ុ, ន, ◌ី)

(3) ស + ◌ុ + ន + ◌ី + ◌ុ + ɪ = (ស, ◌ុ, ន, ◌ី, ◌ុ, ɪ)

---

(4) ស + ួ + ៈ + ៌ + ួ + ឥ = (ស, ួ, ៈ, ៌, ួ, ឥ)

In fact, when there are many possibilities for combining independent typed characters to entry a syllable can cause ambiguity. This leads to more difficulties to match and identify the syllable.
Using the models SC or CC, this phase assign the label for each component cluster in order to convert the syllables into a uniform combination. So, any character appeared in a syllable will have the correct order and the consistent position to match. In the example above, all three combinations (2), (3), (4) have been converted into (1).

*Handling different sign in syllable but identical morpheme*

In Khmer language, there are different signs to use to entry a syllable but identical morpheme which maybe causes the misspelling. For example, there are three typing ways to entry the syllable ក្បៀ containing 5 characters as follow:

(1) ក + ៌ + ួ + ស + ៌ = ក្បៀ = (ក, ៌, ួ, ស, ៌)

(2) ក + ួ + ស + ៌ + ៌ = ក្បៀ = (ក, ួ, ស, ៌, ៌)

(3) ក + ួ + ស + ៌ + ៌ = ក្បៀ = (ក, ួ, ស, ៌, ៌)

The typing way (3) has the same morpheme with (1) and (2) but misspelled. In this phase, (2) was transferred into (1) and replace the character ៌ in (3) into ៌ for the correct spelling.

*Handling different consonant leg sign in syllable but identical morpheme and different meaning*
The combination of the sub-consonant ឥ or ម with the consonant leg sign ួ can form one morpheme but different meanings. For example:

(1)    ួ + ឥ = ្ឥ (subconsonant ឥ)

(2)    ួ + ម = ្ម (subconsonant ម)

The morphemes in (1) with subconsonant ឥ and in (2) with sub-consonant ម are the same. This phase changes sub-consonant signs follow the principle: if the main consonant is កា then the sub-consonant is ម ; conversely, if the main consonant is not ម, use sub-consonant ឥ.

### 3.2.3. Assigning position label for all component clusters

In general, each entry of LDB has the form:
**W = A₁A₂... Aₘ**        where **A**$_i$, i=1..m, is a syllable, m≥1
**A**$_i$ **= C₁C₂ ... C**$_k$ where **C**$_j$, j=1..k, is a component cluster, k≥1

According to CC model, each syllable containing up to m=**3** clusters, each cluster containing up to k≥1 characters (consonants, vowels or signs). There are no marks to identify each syllable A$_i$, i=1..m in the entry W. It is necessary to find the beginning position and the end position of A$_i$ which can contain k clusters Cj, j=1..k, need to exactly locate. Assuming t$_j$ is the label to assign location for each cluster C$_j$. There are 4 types of labels as follow::

LL    Assigned to the left cluster of the syllable; a new syllable can be appear when combined this cluster with its right clusters.

RR    Assigned to the right cluster of the syllable; a new syllable can be appear when combined this cluster with its left clusters.

MM  Assigned to center cluster of syllable.

LR    In case of being independent syllable.

The determination of the position of each component cluster in syllable to assign labels uses three supplementary DB (3), (4) and (5). Here (3) contain all their characteristics. Assuming $(C_j^p)$ specify the characteristic of $C_j$, where $p \in \{-2, -1, 0, 1, 2\}$, there are 5 cases occurred for $C_j$ as follow:

$(C_j^0)$  : $C_j$ is the courrent cluster.
$(C_j^{-1})$ : The left cluster of $C_j$.
$(C_j^{-2})$ : The left cluster of $(C_j^{-1})$.
$(C_j^1)$  : The right cluster of $C_j$.
$(C_j^2)$  : The right cluster of $(C_j^1)$.

So, the labeling for location of component cluster from $(C_j^0)$ can adding 4 possibilities:

$(C_j^{-2}, C_j^{-1}), (C_j^{-1}, C_j^1), (C_j^1, C_j^2), (C_j^{-2} C_j^{-1}, C_j^1 C_j^2)$

The training database (TDB), named (4), contain the selected syllables combined from all overlapping component clusters causing the ambiguity. For example, the syllable ភាគរយ (percent) containing ភាគ and រយ is updated in (4), because this two clusters are overlapping ambiguity and existing in the TDB.

The DB (5) treat the case of consecutive clusters which do not contain characteristic signs to identify syllable. Here (5) contains all the component clusters appearing in DB (2) but not existing in the TDB (4). From the DB (2) containing all the component clusters, it is need to search all items having maximum two clusters satisfying the syllable model SC, or CC. If the item just is a cluster, it is labelled in left and right. After that, this item is updated into labelled cluster DB (6). It is true of the cluster $C^{-1}C^0C^1$ has no distinctive signs to identify syllable. So, from (5), it can be manually labelled $C^1(LL)$, $C^0(RR)$, $C^1(LR)$ or $C^1(LR)$, $C^0(LL)$, $C^1(RR)$ and then updated into (6). The TABLE X. following illustrates the labeling for the component cluster divided from an entry:

TABLE X.       ASSIGNING LOCATION FOR CLUSTERS.

| Entry | វិទ្យាសាស្ត្រកុំព្យូទ័រ (*Information Technology*) | | | | | | | |
|---|---|---|---|---|---|---|---|---|
| Syllables | វិទ្យា | | សាស្ត្រ | | កុំ | ព្យូ | ទ័រ | |
| Clusters | វិ | ទ្យា | សា | ស្ត្រ | កុំ | ព្យូ | ទ័ | រ |
| Labels | LL | RR | LL | RR | LR | LR | LL | RR |
| Positions | 0 | 1 | 2 | 3 | 4 | 5 | 6 | 7 |

### 3.2.4. Identifying the syllable limits

*Matching cluster using the training database*

If the process of labeling find out ambiguous cluster occurs, the identify system match this cluster with data of TDB (4). All any found reasonable clusters are transfered into syllables limits based on the position label of the cluster in syllables. For example, the syllable គមគ (xxx) is divides into three clusters causing ambiguity and can be labelled គ(LL), ម(RR), គ(LR) or គ(LR), ម(LL), គ(RR). If the system cannot found the characteristic signs to identify syllable, use (5) to conduct the matching.

*Combining the position labels*

After all clusters $C_j$ in each each syllable $A_i$ have been labelled the corresponding position, the system combine all the assigned labels to form a complete syllable. And then, this syllable $A_i$ is updated into final Khmer SDB (7). In result display, system put a separator | between two consecutive syllables. Such as the above TABLE X. show that the entry វិញ្ញាសាស្រ្តក្បួនីម៌ (combining 8 clusters) is divided into 5 syllables: វិញ្ញា | សាស្រ្ត | ក្បួ | នី | ម៌ to be updated into (7).

### 3.2.5. Completing the syllable database

At first, the content of (7) consists of all the syllables of (6). The process of updating is performed after the labeling position for all component clusters and the determination of the syllable limits using (5) are done. The TABLE XI. below shows in result the syllables divided from some of the entries in LDB (1) are totally updated into (7).

TABLE XI.        SYLLABLE UDATED RESULT.

| Input | Type | Meaning | Syllables |
|---|---|---|---|
| ទៅ | Simple word | *go* | ទៅ |
| កញ្ញា | Simple word | *girl/teenagers* | កញ្ញា |
| ភាគរយ | Compound word | *percent* | ភាគ\| រយ |
| ស្រុងប្បដាយ | Compound word | *origin* | ស្រុង\| ប្បដាយ |
| កញ្ញាឈើម៉ូតែ | Phrase | *model* | កញ្ញា\| ឈើ \| ម៉ូតែ |
| កណ្ដាលទីក្រុង | Phrase | *city center* | កណ្ដាល \| ទី \| ក្រុង |
| … | | | … |

### 3.3. Assessment

After running the test on our PC system, we invite two Khmer experts who are currently teaching-researchers at the Tra Vinh University to test and assess the results independently. The evaluation is the calculus of percentage (%) of totally correct syllables updated into (7). The TABLE XII. give the assessment results of two experts offering.

TABLE XII.        EVALUATION.

| Type | Quantity | Syllable | Expert 1 | Expert 2 | Average |
|---|---|---|---|---|---|
| Simple word | 7278 | 7278 | 95% | 95% | 95% |
| Compound word | 17095 | 34190 | 93% | 92% | 92.5% |
| Phrase | 24574 | 96779 | 90% | 89% | 89.5% |
| **Total** | **48947** | **138247** | **92.6%** | **92%** | **92.3%** |

The building of SDB (7) using two models SM and CM achieving convinced results. The percentage of automatic division is only correct at 92.3% for these two reasons. First, the LDB (1), some of the entries containing the words borrowed from Pali and Sanskrit do not have the Khmer syllable structure. In order to obtain better results, it is necessary to select all these borrowed words appearing in LDB (1) to correct manually according to the case used. After that, update all these results again into (1). On the other hand, the TDB (4) can not be served to automatically treat all the syllables recently appeared which can occur ambiguity. To solve this problem, it is necessary to increase the size of LDB (1) to extend the frequency of the syllables divided into component clusters in the initial TDB.

## 4. CONCLUSION

The proposal for using two models SC and CC serving to build the SDB is feasible. The solution permit execute the process of the division of an entry from LDB (1) into component clusters stored in (2) with the necessary pretreatment step, labeling for their location using three specific DB from (3) to (5) and storing all the labelled clusters into temporary DB (6). After the final step, identifying the syllable limits from (6), the high precision of test results is really achieved, with the ability to resolve effective the ambiguity. The applying of this solution provide the ability to resolve effective the problem of WS from Khmer documents.

This Khmer SDB is the first result, available, catering for more universal use for different purposes, solving not only WS problem, but also a some of other problems as spelling checking, document classification, document analysis, etc. This is a significant contribution for the process of Khmer language processing.

The directions for further research in this domain is the continuation to improve the solution for the Khmer WS problem with the ability to treat thoroughly ambiguity for all text containing borrowed words, not purely text elements.

## ACKNOWLEDGEMENTS

This paper would not have been done if there were no supports from two experts on Khmer language in Tra Vinh University, Tang Van Thon and Nguyen Ngoc Chau, on checking the test results in our research. Please allow us to show our prondement thankful to these experts.

**Authors**

**Tran Van Nam** – graduated from Can Tho university, majored in Information Technology in 20 00 and earned Master of Computer Science of the University of Da Nang in 2013. I have worked for Tra Vinh University since 2001. I am currently paying attention to natural language processing, data mining, artificial intelligence when I have participated in computer Science from Da Nang university in 2014.

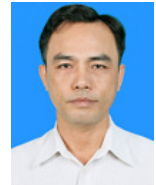

**Nguyen Thi Hue -** graduated from Can Tho university, majored in English Language teaching in 1994 and MA in TESOL of Hanoi Foreign Language University and doctorate degree in Literature of HCM National University in 2011. I have worked for Tra Vinh University since 1994. I am currently interested in study the relationships between  Khmer language with Vietnamese.

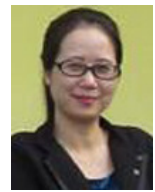

**Phan Huy Khanh** – Graduated from the Hanoi University of Science and Technology (HUST) Vietnam, majored in Calculation Mathematics and Programming, doctorate degree in Computer Science at Lille1 University, Republic of France in 1992 and conferred the title of Associate Professor in 2005. He worked at present at the Faculty of Information Technology, University of Science and Technology (DUT), University of Danang (UDN).He currently teach and research on the domain of Natural Language Processing, Artificial Intelligence, Information Systems. All his articles were stored on Mediafire: https://www.mediafire.com/folder/9b81l9mfnt7xa/BaiBaoKhoaHoc.

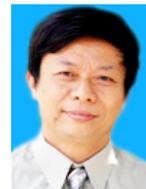